\documentclass[runningheads]{llncs}
\usepackage{graphicx}
\usepackage[english]{babel}
\usepackage[utf8]{inputenc}
\usepackage[caption=false,font=footnotesize]{subfig}
\usepackage{caption}
\usepackage{booktabs}
\usepackage{amsmath,amssymb}
\usepackage{hyperref}
\usepackage[table]{xcolor}
\usepackage{collcell}
\usepackage{hhline}
\usepackage{pgf}
\usepackage{multirow}
\usepackage{hyperref}
\usepackage{blindtext}
\usepackage{enumitem}
\usepackage{wrapfig}
\usepackage{soul}
\usepackage[normalem]{ulem}

\def\colorModel{hsb} %
\newcommand\ColCell[1]{
  \pgfmathparse{#1<50?1:0}  %
    \ifnum\pgfmathresult=0\relax\color{white}\fi
  \pgfmathsetmacro\compA{0}      %
  \pgfmathsetmacro\compB{#1/100} %
  \pgfmathsetmacro\compC{1}      %
  \edef\x{\noexpand\centering\noexpand\cellcolor[\colorModel]{\compA,\compB,\compC}}\x #1
  } 
\newcolumntype{E}{>{\collectcell\ColCell}m{0.45cm}<{\endcollectcell}}  %

\usepackage[misc,geometry]{ifsym}

\newcommand{\srarrow}{$\mbox{\small$\,\rightarrow\,$}$}
\newcommand{\slarrow}{$\mbox{\small$\,\leftarrow\,$}$}

\begin{document}
\title{Balanced-MixUp for Highly Imbalanced \\Medical Image Classification}
\author{Adrian Galdran\inst{1}$^\textrm{,\Letter}$ \and
Gustavo Carneiro\inst{2} \and
Miguel A. González Ballester\inst{3,4}}
\authorrunning{A. Galdran et al.}
\institute{Bournemouth University, UK, \email{agaldran@bournemouth.ac.uk}
\and
University of Adelaide, Adelaide, Australia, \email{gustavo.carneiro@adelaide.edu}
\and
BCN Medtect, Dept. of Information and Communication Technologies, \\Universitat Pompeu Fabra, Barcelona, Spain, \email{ma.gonzalez@upf.edu}
\and
Catalan Institution for Research and Advanced Studies (ICREA), Barcelona, Spain }
\maketitle              %
\begin{abstract}
Highly imbalanced datasets are ubiquitous in medical image  classification problems. In such problems, it is often the case that rare classes associated to less prevalent diseases are severely under-represented in labeled databases, typically resulting in poor performance of machine learning algorithms due to overfitting in the learning process. 
In this paper, we propose a novel mechanism for sampling training data based on the popular MixUp regularization technique, which we refer to as Balanced-MixUp. 
In short, Balanced-MixUp simultaneously performs regular (\textit{i.e.}, instance-based) and balanced (\textit{i.e.}, class-based) sampling of the training data.
The resulting two sets of samples are then mixed-up to create a more balanced training distribution from which a neural network can effectively learn without incurring in heavily under-fitting the minority classes. 
We experiment with a highly imbalanced dataset of retinal images (55K samples, 5 classes) and a long-tail dataset of gastro-intestinal video frames (10K images, 23 classes), using two CNNs of varying representation capabilities. Experimental results demonstrate that applying Balanced-MixUp outperforms other conventional sampling schemes and loss functions specifically designed to deal with imbalanced data. 
Code is released at \url{https://github.com/agaldran/balanced_mixup} .
\end{abstract}

\section{Introduction}
Backed by the emergence of increasingly powerful convolutional neural networks, medical image classification has made remarkable advances over the last years, reaching unprecedented levels of accuracy \cite{litjens_survey_2017}. 
However, due to the inherent difficulty in collecting labeled examples of rare diseases or other unusual instances in a medical context, these models are often trained and tested on large datasets containing a more balanced distribution with respect to image classes than the one found in real-life clinical scenarios, which typically has a long-tailed distribution. 
Under such severe data imbalance, over-represented (i.e., majority) classes tend to dominate the training process, resulting in a decrease of performance in under-represented (i.e., minority) classes \cite{zhuang_care_2019,quellec_automatic_2020}. 
Therefore, developing training methods that are adapted to strong data imbalance is essential for the advance of medical image classification.

Common solutions to address data imbalance involve data re-sampling to achieve a balanced class distribution \cite{buda_systematic_2018}, curriculum learning \cite{jimenez-sanchez_medical-based_2019}, adapted loss functions, \textit{e.g.} cost-sensitive classification \cite{zhou_training_2006,galdran_cost-sensitive_2020}, or weighting the contribution of the different samples \cite{cui_class-balanced_2019,lin_focal_2020}. 
Another approach uses synthetic manipulation of data and/or labels to drive the learning process towards a more suitable solution, like label smoothing \cite{galdran_non-uniform_2020} or SMOTE \cite{chawla_smote_2002}. 
Our proposed technique is a combination of modified training data sampling strategies with synthetic data manipulation via the well-known MixUp regularization method \cite{zhang_mixup_2018}. 
Therefore it is deeply connected with MixUp, with the fundamental difference that, while MixUp randomly combines training samples without taking into account their classes, we carefully mix examples from minority categories with examples from the other classes in order to create a more suitable training data distribution. %

\begin{figure*}[t]
\centering
\subfloat[]{\includegraphics[width = 0.26\textwidth]{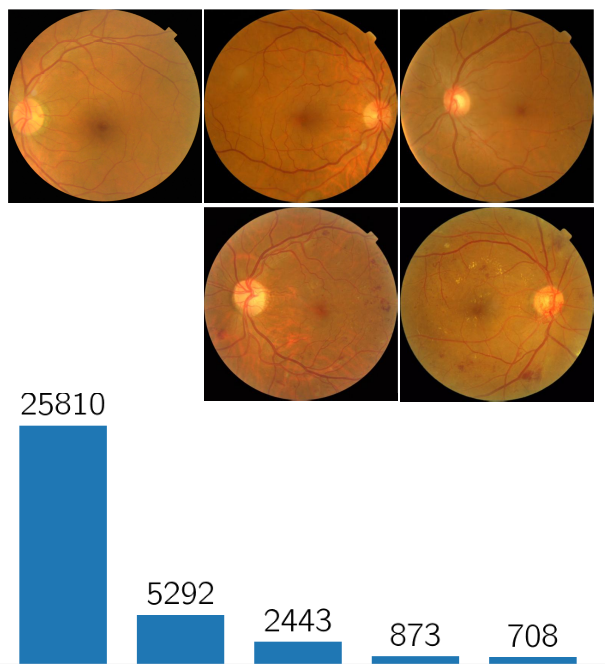}
\label{fig_deg_1}}
\hfil
\subfloat[]{\includegraphics[width = 0.67\textwidth]{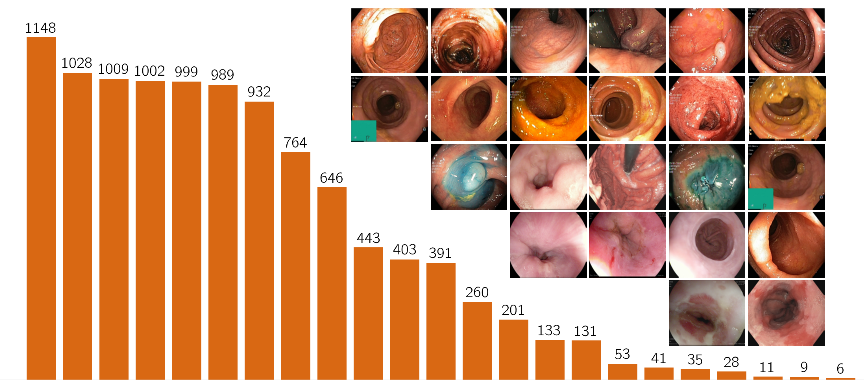}
\label{fig_deg_2}}
\hfil
\caption{The two different data imbalance scenarios considered in this paper. Left: Highly imbalanced problem (DR grading from retinal images, $K=5$ \cite{krause_grader_2018}). Right: Long-tailed data (Gastro-intestinal image classification, $K=23$ \cite{borgli_hyperkvasir_2020}).}
\label{fig_intro}
\vspace{-6.35mm}
\end{figure*}

Some approaches have been recently explored involving MixUp and data imbalance. 
ReMix \cite{chou_remix_2020} mixes up by keeping the minority class label, instead of mixing up the labels.
Similarly, MixBoost \cite{kabra_mixboost_2020} attempts to combine active learning with MixUp to select which training samples to mix from each category, adding an extra complexity layer to the sampling process.
Another popular technique that is related to our approach is SMOTE \cite{chawla_smote_2002}. 
However, SMOTE generates convex combinations of input samples only between nearest neighbors of the same class. 
While this creates extra examples in regions of the space where there is less data, it does not provide the same regularizing benefits as MixUp. 

In this paper, we propose Balanced-MixUp, a new imbalanced-robust training method that mixes up imbalanced (instance-based) and balanced (class-based) sampling of the data.
Experiments on two different medical image classification tasks with highly imbalanced and long-tailed data (as shown in Fig. \ref{fig_intro}), using neural networks of different complexities, show that Balanced-MixUp is effective in creating a more evenly distributed training data and also regularizing overparametrized models, resulting in better performance than other conventional and recent approaches to handle data imbalance in all the considered cases.

\section{Methodology}
In this paper we intend to combine the MixUp regularization technique with a modified sampling strategy for imbalanced datasets. This section introduces each of these concepts and then our proposed combined technique.

\subsection{MixUp Regularization} 
The MixUp technique was initially introduced in \cite{zhang_mixup_2018} as a simple regularization technique to decrease overfitting in deep neural networks. 
If we denote $(x_i, y_i)$ a training example composed of an image $x_i$ and its associated one-hot encoded label $y_i$, with $f_\theta$ being a neural network to be trained to approximate the mapping $f(x_i)=y_i \ \forall i$, then MixUp creates synthetic examples and labels as follows:
\begin{equation}
\hat{x} = \lambda x_i + (1-\lambda) x_j, \ \ \hat{y} = \lambda y_i + (1-\lambda) y_j
\end{equation}
where $\lambda\sim\mathrm{Beta}(\alpha, \alpha)$, with $\alpha>0$. 
Since $\lambda \in [0,1]$, $\hat{x}$ and $\hat{y}$ are random convex combinations of data and label inputs. Typical values of $\alpha$ vary in $[0.1, 0.4]$, which means that in practice $\hat{x}$ will likely be relatively close to either $x_i$ or $x_j$.
Despite its simplicity, it has been shown that optimizing $f_\theta$ on mixed-up data leads to better generalization and improves model calibration \cite{thulasidasan_mixup_2019}.

\subsection{Training Data Sampling Strategies}
When dealing with extremely imbalanced data, the training process is typically impacted by different per-class learning dynamics that result in underfitting of minority classes, which are rarely presented to the model and may end up being entirely ignored \cite{kang_decoupling_2020}. 
Modified sampling strategies can be utilized to mitigate this effect, like oversampling under-represented categories, although merely doing so often leads to counter-productive outcomes, \textit{e.g.} repeatedly showing to the model the same training examples may lead to the overfitting of minority classes \cite{chawla_smote_2002}. 

Before detailing how to combine oversampling of minority classes with MixUp regularization, we introduce some further notation to describe sampling strategies. 
Given a training set $\mathcal{D}=\{(x_i, y_i), \ i=1,...,N\}$ for a multi-class problem with $K$ classes, if each class $k$ contains $n_k$ examples, we have that $\sum_{k=1}^K n_k=N$. We can then describe common data sampling strategies as follows:
\begin{equation}\label{sampling}
p_j = \frac{\displaystyle n_j^q}{\sum_{k=1}^K n_k^q\displaystyle},
\end{equation}
being $p_j$ the probability of sampling from class $j$ during training. 
With this, selecting $q=1$ amounts to picking examples with a probability equal to the frequency of their class in the training set (\textit{instance-based sampling}), whereas choosing $q=0$ leads to a uniform probability $p_j = 1/K$ of sampling from each class, this is, \textit{class-based sampling} or oversampling of minority classes. 
Another popular choice is \textit{square-root sampling}, which stems from selecting $q=1/2$.

\begin{figure*}[t]
\centering
\includegraphics[width=\textwidth]{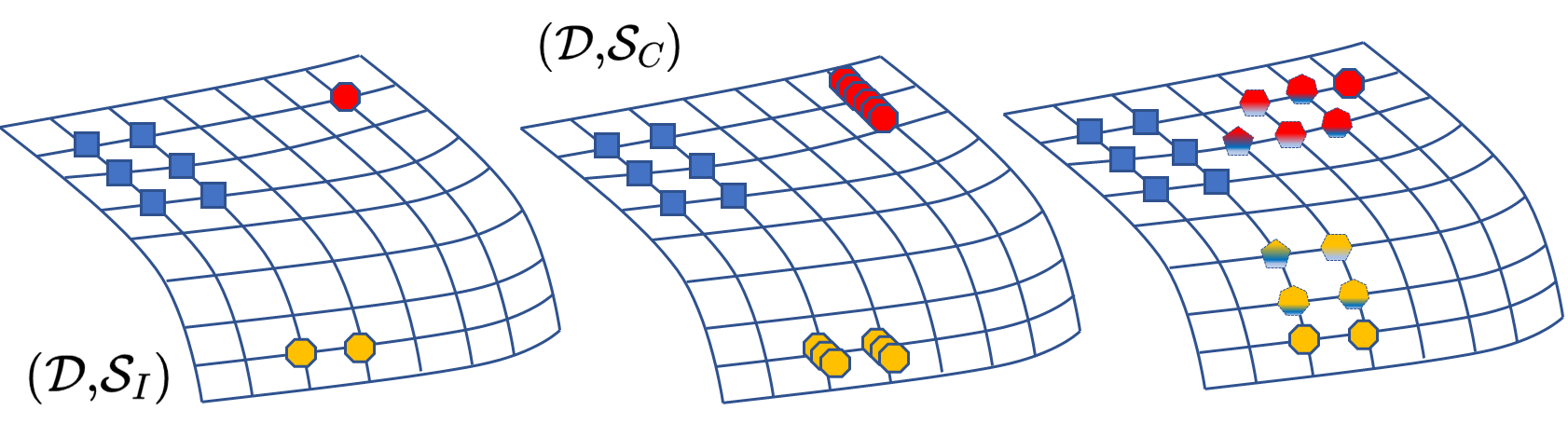}
\caption{Schematic illustration of Balanced-MixUp (right) filling underpopulated areas in the data space - linked to infrequent categories - with synthetic examples.}
\label{fig_scheme}
\end{figure*}

\subsection{Balanced-MixUp}
We propose to combine the two training enhancements described above into a single technique that is suitable for learning under strong data imbalance. 
We introduce an adaptation to this scenario of MixUp, referred to as Balanced-MixUp, in which two data points are not randomly sampled without considering their category, but rather sampling one following an instance-based strategy and the other following a class-based sampling. 

Let us refer to a training dataset together with a sampling strategy as $(\mathcal{D}, \mathcal{S})$, with $\mathcal{S}_I$ and $\mathcal{S}_C$ referring to instance-based and class-based sampling respectively. With this notation, Balanced-MixUp can be described as follows:
\begin{equation}
\hat{x} = \lambda x_I + (1-\lambda) x_C, \ \ \hat{y} = \lambda y_I + (1-\lambda) y_C,
\end{equation}
for $(x_I, y_I) \in (\mathcal{D}, \mathcal{S}_I), (x_C, y_C) \in (\mathcal{D}, \mathcal{S}_C)$. 
This induces a more balanced distribution of training examples by creating synthetic data points around regions of the space where minority classes provide less data density, as illustrated in Fig. \ref{fig_scheme}.
At the same time, adding noise to the labels helps regularizing the learning process.

In our case, departing from the original MixUp formulation, we obtain the mixing coefficient as $\lambda\sim\mathrm{Beta}(\alpha,1)$. This results in an exponential-like distri-
\begin{wrapfigure}{r}{0.55\textwidth}
\vspace{-20pt}
\includegraphics[width=0.49\textwidth]{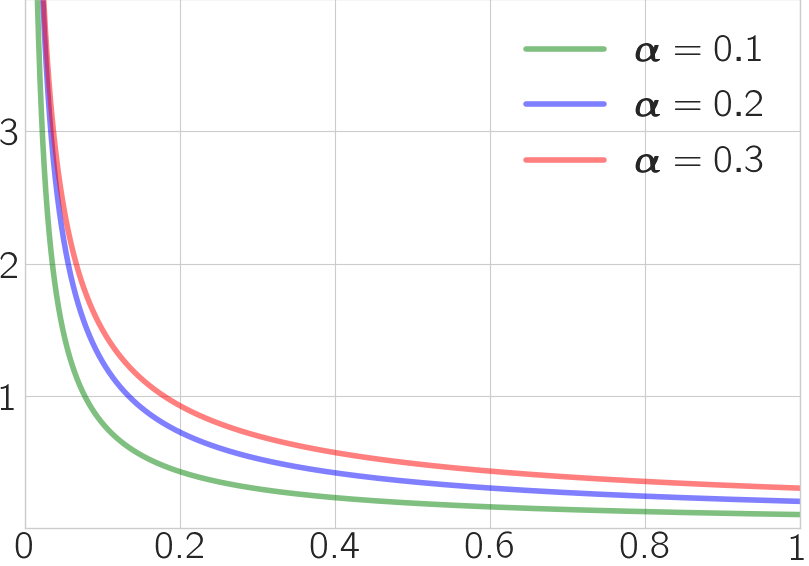}
\caption{$\mathrm{Beta}(\alpha,1)$ distribution for varying $\alpha$}\label{beta}
\end{wrapfigure} 
bution as shown in Fig. \ref{beta}, which leads to convex combinations in which examples from $(\mathcal{D}, \mathcal{S}_I)$ receive more weight, preventing overfitting in minority classes. 
In addition, it allows us to formulate our technique as depending on a single hyperparameter with an intuitive behavior: as $\alpha$ grows, examples from minority classes are mixed up with greater weights, increasing class balance at the expense of risking some overfitting. 

\subsection{Training Details}
All models in this paper are CNNs implemented in Pytorch \cite{paszke_pytorch_nodate} and optimized to reduce the Cross-Entropy loss using Stochastic Gradient Descent with a batch size of $8$ and a learning rate of $0.01$. 
During training, standard data augmentation operations are applied, and the learning rate is cyclically annealed towards 0. 
The metric of interest in each task (see next section) is used to monitor model's performance on the validation set and the best one is kept.
After training, we generate predictions by averaging the result of the input image and its horizontally flipped version, which is common practice to regularize predictions \cite{shanmugam_when_2020}.

\section{Experimental Analysis}
In this section we detail our experimental setup: considered datasets with relevant statistics, baseline techniques for comparison, evaluation metrics, and a discussion on the numerical differences in performance between methods.

\subsection{Experimental Details}
We consider two different image classification tasks: Diabetic Retinopathy (DR) grading from retinal fundus images and gastrointestinal image (GI) classification. 
\begin{itemize}[leftmargin=*]
\item For DR grading, we use the Eyepacs database\footnote{\url{https://www.kaggle.com/c/diabetic-retinopathy-detection}} the largest publicly available dataset for this task. 
This database has images labeled as belonging to one out of five possible DR grades, ranging from DR0 (no DR) to DR4 (proliferative DR). 
Due to the nature of this disease, and the very subtle signs associated to the DR1 class (just a few microaneurysms present), the class distribution is highly imbalanced, as shown in Fig. \ref{fig_intro}. 
The database contains $\sim$35,000 training examples and $\sim$55,000 images assigned to a separate test set. 
We use $10\%$ of the training set for validation and early-stopping purposes, and derive performance measures from the official test set. 
As an external database to examine model generalization we also use the 1748 retinal images in the Messidor-2 dataset\footnote{\url{https://www.adcis.net/en/third-party/messidor2/}}, with the grades provided\footnote{\url{https://www.kaggle.com/google-brain/messidor2-dr-grades}} in \cite{krause_grader_2018}.
\item For GI image classification, we use the Hyper-Kvasir dataset\footnote{\url{https://endotect.com/}}, recently made available \cite{borgli_hyperkvasir_2020}. This database contains 10,662 images labeled with anatomical landmarks as well as pathological and normal findings, and it represents a challenging classification problem due to the larger amount of classes (23) and the varying class frequencies, with minority classes very rarely represented, see Fig. \ref{fig_intro}. 
Please note that there is also an official test set \cite{hicks_endotect_2020}, but we found this to contain only 11 different classes. 
Since our focus is on assessing model performance under a long-tail distribution, we prefer to discard this set, running all our experiments and computing performance for all models in a stratified 5-fold cross-validation manner.
\end{itemize}

For performance assessment in the DR case, we adopt as the main evaluation a measure based on the quadratic-weighted kappa score (quad-$\kappa$), which is the standard performance measure in grading problems, as it considers the correct ordering of predictions.
We also report Matthews Correlation Coefficient (MCC), which has been shown to be meaningful under strong class imbalance \cite{boughorbel_optimal_2017}, and the Kendall-$\tau$ coefficient. 
For the GI image classification task, we adopt as the main assessment measure MCC, as it was used for performance ranking in a recent challenge associated to this database \cite{hicks_endotect_2020}.
Since we are dealing with a long-tail data distribution, we also report Balanced Accuracy (the mean of the diagonal in a normalized confusion matrix) and the macro-F1 score, which capture well performance in minority classes.

For comparison purposes, in both tasks we contrast Balanced-MixUp with other conventional sampling approaches: 1) Class-based sampling (oversampling of minority classes, as recommended in \cite{buda_systematic_2018}), 2) Instance-based sampling, and 3) sampling based on the square root of class-frequencies \cite{mahajan_exploring_2018}. 
We also include training with loss functions that are meant to perform better under severe class imbalance, namely 4) the popular Focal Loss \cite{lin_focal_2017}, which modulates the contribution of each example according to the amount of loss it generates, and 5) a Class-Balanced loss recently introduced that weighs the loss of each class during training based on the \textit{effective} number of samples, see \cite{cui_class-balanced_2019}. 

In all cases, in order to observe the behavior of the different techniques with models of varying sizes, we train two different CNN baseline architectures: a MobileNet V2 \cite{sandler_mobilenetv2_2018} and more powerful ResNext50, modified as in \cite{kolesnikov_big_2020}. 
For sensitivity analysis purposes, we measure the impact of varying the only hyperparameter $\alpha$ in Balanced-MixUp, with $\alpha\in\{0.1, 0.2, 0.3\}$ and report the corresponding results.

\subsection{Discussion on the Numerical Results on DR grading}
Assessment measures for the task of DR grading from retinal images are displayed in Table \ref{tab_results_dr}. 
The leftmost three columns show performance of a MobileNet V2 CNN in terms of quad-$\kappa$, MCC and Kendall-$\tau$. 
In this case, it can be seen that Balanced-MixUp with $\alpha=0.1$ outperforms all other approaches in terms of the three considered figures of merit. 
Note that setting $\alpha=0.2$ still outperforms all other techniques. 
It is also interesting to observe the differences when we switch to a larger model, ResNext50. 
In particular, we can see that: 1) the performance of almost all methods is increased, indicating that ResNext50 is a more suitable choice for this problem; and 2) Balanced-MixUp with $\alpha=0.2$ is now the dominant approach, reaching the highest scores in all considered measures. This is consistent with the intuition that Balanced-MixUp can regularize the learning process, and that increasing its hyperparameter $\alpha$ leads to a stronger regularization
Indeed, for this architecture the second best method is Balanced-MixUp with $\alpha=0.3$. 
It is also worth stressing that, again, the three values of $\alpha$ translate in better results than any of the other considered alternatives.

\begin{table*}[t]  %
	\renewcommand{\arraystretch}{1.3}	
	\centering
\setlength\tabcolsep{4pt}	
\begin{tabular}{l ccc ccc }
& \multicolumn{3}{c}{\textbf{MobileNet V2}} & \multicolumn{3}{c}{\textbf{ResNeXt-50}}  \\
\cmidrule(lr){2-4} \cmidrule(lr){5-7} 
				            &  \textbf{quad-$\kappa$}  &  \textbf{MCC} &  \textbf{Kendall-$\tau$} &  \textbf{quad-$\kappa$}  &  \textbf{MCC} &  \textbf{Kendall-$\tau$}     \\
\midrule
Class-sampling                 				&  74.58  & 49.34 &  63.30 &  74.45  &  51.12  &  64.14    \\
Instance-sampling              				&  78.75  & 61.79 &  73.39 &  80.17  &  62.78  &  74.32    \\
Sqrt-sampling \cite{mahajan_exploring_2018} &  79.32  & 59.66 &  72.22 &  79.38  &  58.77  &  72.22    \\
Focal Loss  \cite{lin_focal_2017}			&  78.59  & 60.74 &  72.70 &  79.73  &  62.66  &  73.37    \\
CB Loss \cite{cui_class-balanced_2019}		&  77.84  & 61.08 &  72.85 &  79.19  &  61.83  &  74.32    \\
\bottomrule
\textbf{Bal-Mxp: $\alpha=0.1$} &  \textbf{79.61}  & \textbf{62.41} &  \textbf{74.05} &  80.35  &  63.31  &  74.74    \\
\textbf{Bal-Mxp: $\alpha=0.2$} &  79.43  & 62.22 &  73.69 &  \textbf{80.78}  &  \textbf{63.78}  &  \textbf{75.05}    \\
\textbf{Bal-Mxp: $\alpha=0.3$} &  78.34  & 61.65 &  73.25 &  80.62  &  63.51  &  74.89    \\
\bottomrule\\[-1.5mm]
\end{tabular}
\caption{Performance comparison with two different CNN architectures for DR grading on the Eyepacs test set. Best results are marked bold.}
\label{tab_results_dr}
\vspace{-10mm}%
\end{table*}%

It is important to stress that the results shown in Table \ref{tab_results_dr} do not represent the state-of-the-art in this dataset.
Recent methods are capable of reaching greater quad-$\kappa$ values by means of specialized architectures with attention mechanisms \cite{he_cabnet_2021}, or semi-supervised learning with segmentation branches \cite{zhou_collaborative_2019}. 
Our results are however aligned with other recently published techniques, as shown in Table \ref{tab_results3}, which also contains test results in the Messidor-2 dataset (without any re-training), demonstrating their competitiveness and generalization ability. 
Moreover we consider Balanced-MixUp to be a supplementary regularizing technique of the learning process from which other methods could also benefit.

\begin{table}[!b]  %
	\renewcommand{\arraystretch}{1.3}	
	\centering
\setlength\tabcolsep{2pt}	
\begin{tabular}{l c c c c c}
& 
\begin{tabular}{@{}c@{}}\textbf{QWKL} \\ \small{PRL'18} \cite{de_la_torre_weighted_2018}\end{tabular}
& 
\begin{tabular}{@{}c@{}} $\mathbf{DR\vert graduate}$ \\ \small{MedIA'20} \cite{araujo_drgraduate_2020} \end{tabular}
&
\begin{tabular}{@{}c@{}} \textbf{Cost-Sensitive} \\ \small{MICCAI'20} \cite{galdran_cost-sensitive_2020} \end{tabular} 
&
\begin{tabular}{@{}c@{}} \textbf{Iter. Aug.} \\ \small{TMI'20} \cite{gonzalez-gonzalo_iterative_2020} \end{tabular} 
&   
\begin{tabular}{@{}c@{}} \textbf{Bal-Mxp} \\ Rx50, $\alpha=0.2$\end{tabular}\\
\cmidrule(lr{.0005em}){1-6} 
\textbf{Eyepacs}               & 74.00  & 74.00  &  78.71 & 80.00  & \textbf{80.78} \\
\textbf{Messidor-2}            & -----  & 71.00  &  79.79 & -----  & \textbf{85.14} \\[0.25cm]
\end{tabular}
\caption{Performance comparison in terms of quad-$\kappa$ for methods trained on the Eyepacs dataset and tested (no retraining) on Eyepacs and Messidor-2.}
\label{tab_results3}
\vspace{-10mm}
\end{table}

\subsection{Discussion on the Numerical Results on GI image classification}
The second task we consider is Gastro-Intestinal image classification. 
Here the class imbalance is greater than in the DR problem, in addition to the presence of 23 categories, which turn this into a more challenging dataset. 
Table \ref{tab_results_gi} shows the median performance in a stratified 5-fold cross-validation process, when considering exactly the same techniques and architectures as in the previous section.

Although in this task the performance measures appear to be more mixed than in the DR problem, we can still see how Balanced-MixUp appears to bring improvements over the other competing approaches. 
Again, the ResNext50 architecture delivers better performance in terms of MCC, although differences are not as high as before, and some methods show a degraded performance when applied with this larger architecture (compared to using MobileNet). 

Focusing first on the MCC score, we can appreciate again the regularizing effect of Balanced-MixUp: for both architectures, applying it leads to improved performance. 
Moreover, we can again see the same trend as above: the MobileNet model performs better with less regularization ($\alpha=0.1$), and increasing $\alpha$ slowly deteriorates the MCC. 
As expected, when using the ResNeXt architecture, a greater $\alpha=0.2$ delivers better performance. 
It is worth noting that the MCC reached by the MobileNet model with $\alpha=0.1$ is greater than in any of the considered methods and architectures, excluding models regularized with Balanced-MixUp and varying $\alpha$. 

Finally, it is relevant to observe that when we consider balanced accuracy and macro-F1, which can capture better performance in minority classes, for almost all methods it appears to be beneficial to use a smaller architecture. This appears to reduce to some extent the minority class overfitting due to oversampling, observed in the low performance of class-based sampling.
Interestingly, even with this small architecture, Balanced MixUp increases the performance for all values of $\alpha$, which further shows that our approach not only helps regularizing large models but also contributes to an improved learning on minority classes. 

\begin{table*}[t]  %
	\renewcommand{\arraystretch}{1.3}	
	\centering
\setlength\tabcolsep{4pt}	
\begin{tabular}{l ccc ccc }
& \multicolumn{3}{c}{\textbf{MobileNet V2}} & \multicolumn{3}{c}{\textbf{ResNeXt-50}}  \\
\cmidrule(lr){2-4} \cmidrule(lr){5-7} 
				               &  \textbf{MCC}  &  \textbf{B-ACC} &  \textbf{Macro-F1} &  \textbf{MCC}  &  \textbf{B-ACC} &  \textbf{Macro-F1}    \\
\midrule
Class-sampling                 				&  89.74  & 61.84 &  61.84 &  88.97  &  59.57  &  58.40    \\
Instance-sampling              				&  90.49  & 62.13 &  62.16 &  90.74  &  61.41  &  61.69    \\
Sqrt-sampling \cite{mahajan_exploring_2018} &  90.11  & 63.01 &  62.66 &  90.29  &  62.84  &  62.85    \\
Focal Loss  \cite{lin_focal_2017}			&  90.30  & 62.09 &  61.78 &  90.23  &  62.12  &  62.36    \\
CB Loss \cite{cui_class-balanced_2019}		&  85.84  & 54.93 &  54.87 &  89.67  &  \textbf{63.84}  &  63.71    \\
\bottomrule
\textbf{Bal-Mxp: $\alpha=0.1$} &  \textbf{90.90}  & 63.49 &  62.77 &  91.05  &  62.55  &  62.92    \\
\textbf{Bal-Mxp: $\alpha=0.2$} &  90.54  & 63.44 &  63.85 &  \textbf{91.15}  &  62.80  &  \textbf{64.00}    \\
\textbf{Bal-Mxp: $\alpha=0.3$} &  90.39  & \textbf{64.76} &  \textbf{64.07} &  90.84  &  62.34  &  62.35    \\
\bottomrule\\[-1.5mm]
\end{tabular}
\caption{Stratified 5-Fold cross-validation (median) results for GI image classification on the Hyper-Kvasir dataset. Best results are marked bold.}
\label{tab_results_gi}
\vspace{-9mm}
\end{table*}%

\section{Conclusion}
Imbalanced data is present in almost any medical image analysis task. 
Therefore, designing techniques for learning in such regimes is a challenge of great clinical significance. 
This paper introduces Balanced-MixUp to deal with heavy imbalanced data distributions by means of the combination of the popular MixUp regularization technique and modified training data sampling strategies. 
Balanced-MixUp is easy to implement and delivers consistent performance improvements when compared with other popular techniques. Its extension to other tasks beyond image classification represents a promising future research direction.

\section*{Acknowledgments}
This work was partially supported by a Marie Skłodowska-Curie Global Fellowship (No 892297) and by Australian Research Council grants (DP180103232 and FT190100525).

\bibliographystyle{splncs04}
\bibliography{miccai_rebalanced_mxp.bib}

\appendix
\section{Appendix: Extended Numerical Results}
\subsection{Numerical Results on DR grading}
For deriving confidence intervals, labels and model predictions on the Eyepacs test set are bootstrapped (n=1000) in a stratified manner according to the presence of DR grades.
Tables \ref{tab_results_dr_mob} and \ref{tab_results_dr_rx50} show the mean/std resulting from this. 

\begin{table*}[!b]  %
	\renewcommand{\arraystretch}{1.02}	
	\centering
\setlength\tabcolsep{12pt}	
\begin{tabular}{l ccc }
& \multicolumn{3}{c}{\textbf{MobileNet V2}}  \\
\cmidrule(lr){2-4}
				            &  \textbf{quad-$\kappa$}  &  \textbf{MCC} &  \textbf{Kendall-$\tau$}     \\
\midrule
Class-sampling                 				&  74.57$\, \pm\, 0.29$  & 49.33$\, \pm\, 0.34$ &  63.29$\, \pm\, 0.31$    \\
Instance-sampling              				&  78.74$\, \pm\, 0.30$  & 61.79$\, \pm\, 0.32$ &  73.38$\, \pm\, 0.28$     \\
Sqrt-sampling  								&  79.32$\, \pm\, 0.00$  & 59.66$\, \pm\, 0.35$ &  72.21$\, \pm\, 0.31$     \\
Focal Loss  								&  78.58$\, \pm\, 0.30$  & 60.74$\, \pm\, 0.34$ &  72.68$\, \pm\, 0.34$    \\
CB Loss 									&  77.83$\, \pm\, 0.32$  & 61.07$\, \pm\, 0.33$ &  72.84$\, \pm\, 0.33$     \\
\bottomrule
\textbf{Bal-Mxp: $\alpha=0.1$} &  $\mathbf{79.61\pm0.30}$  & $\mathbf{62.41 \, \pm\, 0.33}$ &  $\mathbf{74.05\, \pm\, 0.29}$     \\
\textbf{Bal-Mxp: $\alpha=0.2$} &  79.42$\, \pm\, 0.31$  & 62.23$\, \pm\, 0.33$ &  73.68$\, \pm\, 0.29$     \\
\textbf{Bal-Mxp: $\alpha=0.3$} &  78.34$\, \pm\, 0.32$  & 61.65$\, \pm\, 0.34$ &  73.24$\, \pm\, 0.29$    \\
\bottomrule\\[-1.5mm]
\end{tabular}
\caption{Performance comparison on Eyepacs. Best results marked bold.}
\label{tab_results_dr_mob}
\vspace{-5mm}%
\end{table*}%

\begin{table*}[!b]  %
	\renewcommand{\arraystretch}{1.02}	
	\centering
\setlength\tabcolsep{12pt}	
\begin{tabular}{l ccc }
& \multicolumn{3}{c}{\textbf{ResNeXt-50}}  \\
\cmidrule(lr){2-4}
				            &  \textbf{quad-$\kappa$}  &  \textbf{MCC} &  \textbf{Kendall-$\tau$}     \\
\midrule
Class-sampling                 				&  74.45$\, \pm\, 0.30$  & 51.13$\, \pm\, 0.32$ &  64.14$\, \pm\, 0.30$    \\
Instance-sampling              				&  80.17$\, \pm\, 0.30$  & 62.78$\, \pm\, 0.32$ &  74.31$\, \pm\, 0.29$     \\
Sqrt-sampling  								&  79.39$\, \pm\, 0.29$  & 58.78$\, \pm\, 0.33$ &  72.22$\, \pm\, 0.30$     \\
Focal Loss  								&  79.73$\, \pm\, 0.29$  & 62.67$\, \pm\, 0.32$ &  74.32$\, \pm\, 0.28$    \\
CB Loss 									&  79.18$\, \pm\, 0.30$  & 61.84$\, \pm\, 0.32$ &  73.36$\, \pm\, 0.28$     \\
\bottomrule
\textbf{Bal-Mxp: $\alpha=0.1$} 				&  80.35$\, \pm\, 0.29$  & 63.30$\, \pm\, 0.32$ &  74.74$\, \pm\, 0.28$     \\
\textbf{Bal-Mxp: $\alpha=0.2$} 				&  $\mathbf{80.78\, \pm\, 0.27}$  & $\mathbf{63.78\, \pm\, 0.31}$ &  $\mathbf{75.05\, \pm\, 0.27}$     \\
\textbf{Bal-Mxp: $\alpha=0.3$} 				&  80.62$\, \pm\, 0.29$  & 63.50$\, \pm\, 0.32$ &  74.89$\, \pm\, 0.29$     \\
\bottomrule\\[-1.5mm]
\end{tabular}
\caption{Performance comparison on Eyepacs. Best results marked bold.}
\label{tab_results_dr_rx50}
\vspace{-8mm}
\end{table*}%

\subsection{Numerical Results on GI image classification}
Results on GI image classification in our paper were obtained by a stratified 5-fold cross-validation process, due to the absence of data from minority classes in the official test set. 
In the paper we reported the median performance across folds, but it is also useful to observe performance dispersion. 
Tables \ref{tab_results_gi_mob} and \ref{tab_results_gi_rx50} show results in terms of the minimum, median, and maximum performance across folds for each model, displayed as minimum \slarrow median \srarrow maximum values.

\begin{table*}[h]  %
	\renewcommand{\arraystretch}{1.02}	
	\centering
\setlength\tabcolsep{2pt}
\vspace{-4mm}	
\begin{tabular}{l ccc }
& \multicolumn{3}{c}{\textbf{MobileNet V2}}   \\
\cmidrule(lr){2-4}  
				               				&  \textbf{MCC}  &  \textbf{B-ACC} &  \textbf{Macro-F1}     \\
\midrule
Class-sampling                 				&  89.22\slarrow89.74\srarrow90.23  & 61.63\slarrow89.84\srarrow63.13 &  61.14\slarrow61.84\srarrow66.42 \\
Instance-sampling              				&  \underline{90.34}\slarrow90.49\srarrow\underline{91.47}  & 60.92\slarrow62.13\srarrow64.73 &  60.02\slarrow62.16\srarrow64.23     \\
Sqrt-sampling  								&  89.88\slarrow90.11\srarrow90.96  & 61.10\slarrow63.01\srarrow65.49 &  61.37\slarrow62.66\srarrow\underline{71.28}     \\
Focal Loss 									&  90.14\slarrow90.30\srarrow90.60  & 61.55\slarrow62.09\srarrow64.70 &  61.03\slarrow61.78\srarrow64.45     \\
CB Loss 									&  84.09\slarrow85.84\srarrow86.33  & 53.50\slarrow54.93\srarrow56.74 &  53.43\slarrow54.87\srarrow56.01     \\
\bottomrule
\textbf{Bal-Mxp $\alpha=0.1$} 				&  90.14\slarrow\underline{\textbf{90.90}}\srarrow91.04 & 61.57\slarrow63.49\srarrow69.60 			&  61.54\slarrow62.77\srarrow66.48     \\
\textbf{Bal-Mxp $\alpha=0.2$} 				&  90.01\slarrow90.54\srarrow90.91  		   & 60.86\slarrow63.44\srarrow69.85 			&  60.40\slarrow63.85\srarrow68.50     \\
\textbf{Bal-Mxp $\alpha=0.3$} 				&  89.85\slarrow90.39\srarrow90.89  		   & \underline{62.10}\slarrow\underline{\textbf{64.76}}\srarrow\underline{69.89}    &  \underline{61.55}\slarrow\underline{\textbf{64.07}}\srarrow69.48     \\
\bottomrule\\[-1.5mm]
\end{tabular}
\caption{Stratified 5-Fold cross-validation (min, median, max) results for GI image classification on the Hyper-Kvasir dataset. Best results are marked bold.}
\label{tab_results_gi_mob}
\vspace{-15mm}
\end{table*}%

\begin{table*}[h]  %
	\renewcommand{\arraystretch}{1.02}	
	\centering
\setlength\tabcolsep{2pt}	
\begin{tabular}{l ccc }
& \multicolumn{3}{c}{\textbf{ResNeXt-50}}   \\
\cmidrule(lr){2-4} 
				               				&  \textbf{MCC}  &  \textbf{B-ACC} &  \textbf{Macro-F1}  \\
\midrule
Class-sampling                 				&  88.97\slarrow89.84\srarrow90.28  & 59.57\slarrow61.38\srarrow64.05 &  58.40\slarrow61.52\srarrow64.14     \\
Instance-sampling              				&  \underline{90.60}\slarrow90.74\srarrow91.10  & 60.05\slarrow61.41\srarrow62.08 &  59.72\slarrow61.69\srarrow62.32     \\
Sqrt-sampling 								&  89.78\slarrow90.29\srarrow90.49  & 60.70\slarrow62.84\srarrow66.71 &  60.39\slarrow62.85\srarrow66.21     \\
Focal Loss									&  89.93\slarrow90.23\srarrow\underline{91.51}  & 60.23\slarrow62.12\srarrow64.08 &  60.29\slarrow62.36\srarrow64.78   \\
CB Loss 									&  89.63\slarrow89.67\srarrow90.29  & \underline{61.99}\slarrow\underline{\textbf{63.84}}\srarrow\underline{68.81} &  \underline{62.22}\slarrow63.71\srarrow\underline{68.66}    \\
\bottomrule
\textbf{Bal-Mxp $\alpha=0.1$}				&  90.38\slarrow91.05\srarrow91.20  & 60.44\slarrow62.55\srarrow64.37 &  60.90\slarrow62.92\srarrow64.52    \\
\textbf{Bal-Mxp $\alpha=0.2$} 				&  90.59\slarrow\underline{\textbf{91.15}}\srarrow91.20  & 60.70\slarrow62.80\srarrow65.32 &  61.35\slarrow\underline{\textbf{64.00}}\srarrow65.86     \\
\textbf{Bal-Mxp $\alpha=0.3$} 				&  90.49\slarrow90.84\srarrow91.30  & 60.14\slarrow62.34\srarrow65.89 &  60.34\slarrow62.35\srarrow65.62    \\
\bottomrule\\[-1.5mm]
\end{tabular}
\caption{Stratified 5-Fold cross-validation (min, median, max) results for GI image classification on the Hyper-Kvasir dataset. Best results are marked bold.}
\label{tab_results_gi_rx50}
\vspace{-3mm}
\end{table*}%

Due to the few instances in the minority classes, results for the GI classification task are noisier than for the DR grading problem. 
However, one can see (as pointed out in the paper) that for metrics that highlight performance on the minority classes, like Balanced Accuracy or Macro-F1, not only are the overall best median results obtained by the MobileNet model with Balanced-MixUp regularization (with $\alpha=0.3$), but also the best minimum and maximum values are attained by this combination, although it is worth noting the good performance of the CB-loss in conjunction with the ResNeXt50 architecture.

\end{document}